%% file: arxiv_tacl.tex
\documentclass[11pt,a4paper]{article}
\usepackage{times,latexsym}
\usepackage{url}
\usepackage[T1]{fontenc}
\usepackage{pgfplots}
\usepackage{adjustbox}
\usepackage{physics}

\usepackage[acceptedWithA]{tacl2018v2}
\usepackage{xspace,mfirstuc,tabulary}

\newif\iftaclinstructions
\taclinstructionsfalse 
\iftaclinstructions
\renewcommand{\confidential}{}
\renewcommand{\anonsubtext}{(No author info supplied here, for consistency with
TACL-submission anonymization requirements)}
\newcommand{\instr}
\fi

\iftaclpubformat

\else

\fi

\newcommand{\vo}{\vec{o}\@ifnextchar{^}{\,}{}}

\usepackage{mathtools}
\usepackage{booktabs} 
\usepackage{tikz}
\usepackage{multirow}
\usepackage{comment}
\usepackage[normalem]{ulem}
\usepackage{amsmath}
\usepackage{amsfonts}
\usepackage{enumitem}
\usepackage[linesnumbered,algoruled,boxed,noend]{algorithm2e}
\usepackage{xcolor}
\usepackage{color, colortbl}
\usepackage{multirow,booktabs, hhline}
\usepackage{makecell}
\usepackage{caption}
\usepackage{subcaption}
\input{macros}

\def\knn{$k$NN\xspace}
\def\knnplus{$k$NN$^+$\xspace}
\def\deepbetwoknn{BE \knnplus}
\def\deepcetwoknn{CE \knnplus}
\def\modelxlmr{$\mathcal{M}_{feature}^{\text{XLM-R}}$\xspace}
\def\modelpxlmr{$\mathcal{M}_{feature}^{{\text{P-XLM-R}}}$\xspace}
\def\jigsawen{Jigsaw English\xspace}
\def\jigsawm{Jigsaw Multilingual\xspace}
\def\wul{WUL\xspace}

\title{A Neighbourhood Framework for Resource-Lean Content Flagging}

\author{
Sheikh Muhammad Sarwar$^{1,2}$\and
Dimitrina Zlatkova$^1$\and
Momchil Hardalov$^{1,3}$\\\and
\textbf{Yoan Dinkov}$^1$\and
\textbf{Isabelle Augenstein}$^{1,4}$\and
\textbf{Preslav Nakov}$^{1,5}$\\
$^1$Checkstep Research,
$^2$University of Massachusetts, Amherst,\\
$^3$Sofia University,
$^4$University of Copenhagen,
$^5$Qatar Computing Research Institute, HBKU,\\
smsarwar@cs.umass.edu,\\ \{didi, momchil, yoan.dinkov, isabelle, preslav.nakov\}@checkstep.com
}

\date{}

\begin{document}
\maketitle

\begin{abstract}

We propose a novel framework for cross-lingual content flagging with limited target-language data, which significantly outperforms prior work in terms of predictive performance. The framework is based on a nearest-neighbour architecture. It is a modern instantiation of the vanilla $k$-nearest neighbour model, as we use Transformer representations in all its components. Our framework can adapt to new source-language instances, without the need to be retrained from scratch. Unlike prior work on neighbourhood-based approaches, we encode the neighbourhood information based on query--neighbour interactions. We propose two encoding schemes and we show their effectiveness using both qualitative and quantitative analysis. Our evaluation results on eight languages from two different datasets for abusive language detection show sizable improvements of up to 9.5 F1 points absolute (for Italian) over strong baselines. On average, we achieve 3.6 absolute F1 points of improvement for the three languages in the Jigsaw Multilingual dataset and 2.14 points for the WUL dataset.
\end{abstract}

\input{introduction}
\input{related_work}
\input{proposed_model}

\section{Experimental Setting}

\subsection{Datasets}

We conducted experiments on two different multilingual datasets covering eight languages from six language families: Slavic, Turkic, Romance, Germanic, Albanian, and Finno-Ugric. We used these datasets as our target datasets, and an English dataset as the source dataset, which contains a large number of training examples with fine-grained categorisation. Both the source and target datasets are from the same domain (Wikipedia), as we do not study domain adaptation techniques in the present work. We describe these three datasets in the following paragraphs. The number of examples per dataset and the corresponding label distributions are shown in Table~\ref{tab:label-distribution}.

\paragraph{Jigsaw English}~\cite{jigsaw} is an English dataset with over 159K manually reviewed comments, annotated with multiple labels. We map the labels (\emph{toxic}, \emph{severe toxic}, \emph{obscene}, \emph{threat}, \emph{insult}, and \emph{identity hate}) into a \emph{flagged} label; if at least one of these six labels is present for some example, we consider that example as \emph{flagged}, and as \emph{neutral} otherwise. As Jigsaw English is a resource-rich dataset, covering different aspects of abusive language, we use it as the source dataset. We use all its examples for training, as we validate our models on the \emph{target} datasets' dev sets. 

\paragraph{Jigsaw Multilingual}~\cite{jigsaw-multilingual} aims to improve toxicity detection by addressing the shortcomings of the monolingual setup. The dataset contains examples in Italian, Turkish, and Spanish. It has binary labels (toxic or non-toxic), and thus it aligns well with our experimental setup. The label distribution is fairly similar to that for \jigsawen, as shown in Table~\ref{tab:label-distribution}. This dataset is used for experimenting in a resource-rich environment. As it does not have standard training, testing, and development sets, we split the examples in each language as follows: 1,500, 500, and 500 for Italian and Spanish, and 1,800, 600, and 600 for Turkish.

\paragraph{WUL} \cite{glavas-etal-2020-xhate} aims to create a fair evaluation setup for abusive language detection in multiple languages. Although originally in English, multilinguality is achieved by translating the comments as accurately as possible into five different languages: German (DE), Hungarian (HR), Albanian (SQ), Turkish (TR), and Russian (RU).

We use this dataset partially, by using the test set originally generated by \citet{wulczyn2017ex}, who focused on identifying personal attacks. In contrast to \jigsawm, this dataset is used for experimenting in a low-resource environment. For each language, we have 600 examples, which are split as follows: 400, for training, 100 for development, and 100 for testing. As abusive content can be very culture-specific, there will be cases, even within the same language, where some utterances will be offensive in one culture, but not in another one. Thus, a translation-based dataset such as \wul{} might not be an ideal choice, and we acknowledge this limitation.

The results from experimenting with the above datasets cannot be compared to those in the literature as we use the test set from these datasets to create our train/dev/test splits. The datasets used in previous work (Jigsaw Multilingual and WUL) provide English-only training data and observe the performance of different models in zero-shot transfer learning settings. Our setup is different as we assume that there is a limited number of training examples in the target language. Thus, we produce results only on a subset of the original testset for both datasets. Therefore, our results are not directly comparable to the results from the literature, as both the training and the testing datasets differ.

\begin{table}
    \centering
\resizebox{0.49\textwidth}{!}{
    \begin{tabular}{lrcc}
        \toprule
        \bf{Dataset} & \bf {Examples} & \bf{Flagged \%} & \bf{Neutral \%}\\
         \midrule
         Jigsaw En & 159,571 & 10.2 & 89.8 \\
         \midrule
         Jigsaw Multi & 8,000 & 15.0 & 85.0  \\
         WUL & 600 & 50.3 & 49.7 \\
         \bottomrule
    \end{tabular}
}
    \caption{Statistics about the dataset sizes and the respective label distributions.}
    \label{tab:label-distribution}
\end{table}

\subsection{Baselines} 
We compare our proposed approach against three families of strong baselines. The first one considers training models only on the target dataset, the second one is source adaptation, where we use \jigsawen as our source dataset, and the third one consists of traditional \knn{}\xspace classification method, but with dense vector retrieval using LaBSE~\cite{feng2020language}. We use cosine similarity under a LaBSE representation space to retrieve neighbours for the baselines and for our proposed approaches.

\paragraph{\underline{Target Dataset Training}} This family of baselines uses only the target dataset for training:

\textbf{Lexicon} approach: After standard text tokenization and normalization of the text, we count the number of terms it contains that are also listed in the abusive language lexicon
 HurtLex\footnote{\url{https://github.com/valeriobasile/hurtlex}}. Based on the development set, we learn a threshold for the minimum number of matches required to flag the text. Then, we apply the lexicon and the threshold to the test set.

\textbf{fastText} is a baseline that uses the mean of the token vectors obtained from fastText~\citep{joulin-etal-2017-bag} word embeddings to represent a textual example. These representations are then used in a binary logistic regression classifier.

\textbf{XLM-R Target} is a pre-trained XLM-R model, which we fine-tune on the target dataset. 

\input{tables/overall_results}

\paragraph{\underline{Source Adaptation}} This family of baselines includes variations of XLM-R:

\textbf{XLM-R Mix-Adapt} is a baseline model, which we train by mixing source and target data. This is possible because the label inventories of our source and target datasets are the same: $\mathcal{Y} = \{flagged, neutral\}$. The mixing is done by oversampling the target data to match the number of instances of the source dataset. As the number of instances in the target dataset is limited, this is preferable to undersampling.  
    
\textbf{XLM-R Seq-Adapt} \cite{garg2020tanda} is a Transformer pre-trained on the source and fine-tuned on the target data. Here, we fine-tune XLM-R on the \jigsawen dataset, and then we do a second round of fine-tuning on the target dataset. 

\paragraph{\underline{Nearest Neighbour}} We apply two nearest neighbour baselines, using majority voting for label aggregation. We varied the number of neighbours from 3 to 20, and we found that using 10 neighbours works best (on the dev set).  

\textbf{LaBSE-kNN} Here the source dataset is indexed using representations obtained from LaBSE sentence embeddings~\cite{feng2020language}, and the neighbours are retrieved using cosine similarity.
    
\textbf{Weighted LaBSE-kNN} is a baseline that uses the same retrieval step as LaBSE-kNN, but with a weighted voting strategy: each label is scored by summing the cosine similarities for the retrieved flagged and neutral neighbours, respectively; then, the label with the highest score is returned.

\subsection{Evaluation Measures}

Following prior work on abusive language detection, we use F1 measure for evaluation. The F1 measure combines precision and recall (using a harmonic mean), which are both important to consider for automatic abusive language detection systems. In particular, online platforms strive to remove all content that violates their policies, and thus, if the system were to achieve 100\% recall, the contents could be further filtered by human moderators to weed out the benign content. However, if the system's precision were very low, it would mean that the moderators would have to read every piece of content on the platform.

\subsection{Fine-Tuning and Hyper-Parameters}
\label{sec:finetune}

We train all the models for 10 epochs with XLM-R as a base transformer representation with a maximum sequence length of 256 tokens. However, we make an exception for SRC (see Section~\ref{sec:overall_results}): we train it for a single epoch, 
as training a neighbourhood-based model on a large dataset is resource-intensive. For all the approaches, we use Adam with $\beta_1$ 0.9, $\beta_2$ 0.999, $\epsilon$ 1e-08 as the optimiser setting. For the baseline models, we use a batch size of 64, and a learning rate of 4e-05. For \knnplus-based models, we create a training batch from a query and its 10-nearest neighbours. 
For stable updates, we accumulate gradients from 50 batches before back-propagation. We selected the values of all of the aforementioned hyper-parameters based on the validation set. For \knnplus-based models, the best learning rate is selected from \{5e-05, 7e-05\}.

\section{Experimental Results}

\subsection{Evaluation in a Cross-lingual Setting}
\label{sec:overall_results}

 Table~\ref{tab:overall} shows the performance of our model variants compared to the seven strong baselines we described above (rows 1--7). The first two rows represent non-contextual baselines and they perform worse compared to the baseline pre-trained XLM-R models fine-tuned with labelled data (rows 3--5). Specifically, the lexicon baseline performs the worst among all, which indicates the limited coverage of hate speech lexicon and the loss in precision due to token mismatches and context obliviousness. For example, the word \emph{monkey} is generally included in a hate speech lexicon, but the appearance of the token in a textual content does not necessarily mean that the content is abusive.

The {\setlength{\fboxsep}{0pt}\colorbox{Red!60}{\strut high}\colorbox{Blue!60}{\strut{}lighted}} rows in Table~\ref{tab:overall} show different variants of our framework, based on \deepcetwoknn{} and \deepbetwoknn{}, i.e., using cross-encoders vs. bi-encoders. For each of the encoding schemes, we instantiate three different models by using three different pre-trained representations fine-tuned in our neighbourhood framework, namely: \modelxlmr, which is a pre-trained XLM-RoBERTa model (XLM-R); \modelpxlmr, which is an XLM-R model fine-tuned under a knowledge distillation setting with 50 million paraphrases and parallel data in 50 languages~\cite{reimers-gurevych-2020-making}; and \modelpxlmr $\rightarrow$ SRC, which is an \modelpxlmr model fine-tuned with source data (here, 159,571 instances from \jigsawen) in our neighbourhood framework.

In order to train with SRC, we use all the training data in \jigsawen, and we retrieve neighbours from \jigsawen using LaBSE sentence embeddings.\footnote{Note that we only use LaBSE for retrieval, as it has a large coverage of languages.}
Then, we use this training data to fine-tune \modelpxlmr with our \knnplus{}-based cross-encoder (\deepcetwoknn{} + \modelpxlmr $\rightarrow$ SRC) and bi-encoder (\deepbetwoknn{} + \modelpxlmr $\rightarrow$ SRC) experimental setups.

This is analogous to applying sequential adaptation \cite{garg2020tanda}, but here we do it in our neighbourhood framework. 

The SRC approach addresses one of the weaknesses of our \knn{} framework. The training data is created from instances in the target dataset and their neighbours from the source dataset. Thus, the neighbourhood model cannot use all source training data, as it pre-selects a subset of the source data based on similarity. This is a disadvantage compared to the sequential adaptation model, which uses all source training instances for pre-training. In order to overcome this, we use the neighbourhood approach to pre-train our models with source data.

Table~\ref{tab:overall} shows the F1 scores for eight language-specific training and evaluation sets stemming from two different data sets: \jigsawm and \wul{}. \jigsawm is an imbalanced dataset with 15\% abusive content and \wul{} is balanced (see Table~\ref{tab:label-distribution}). Thus, it is hard to achieve high F1 score in \jigsawm, whereas for \wul{} the F1 scores are relatively higher. Our \deepcetwoknn{} variants achieve superior performance to all the baselines and our \deepbetwoknn{} variants as well in the majority of the cases. 

The performance of the best and of the second-best models for each language are highlighted by \textbf{bold-facing} and \underline{underlining}, respectively. We attribute the higher scores achieved by \deepcetwoknn{} variants compared to the \deepbetwoknn{} on the late-stage interaction of the query and its neighbours. 

The \deepcetwoknn{} variants show a large performance gain compared to baseline models on the Italian and the Turkish test sets from \jigsawm. Even though the additional SRC pre-training is not always helpful for the \deepcetwoknn{} model, it is always helpful for the \deepbetwoknn{} model. However, both models struggle to outperform the baseline for the Spanish test set. We analysed the training data distribution for Spanish, but we could not find any noticeable patterns.

Yet, it can be observed that the XLM-R target baseline for Spanish (2nd row, 1st column) achieves a higher F1 score compared to the Seq-Adapt baseline, which yields better performance for Italian and Turkish. We believe that the in-domain training examples are good enough to achieve a reasonable performance for Spanish. 

On the \wul{} dataset, \deepbetwoknn{} + \modelpxlmr with SRC pre-training outperforms the \deepcetwoknn{} variants and all baselines for Albanian, Russian, and Turkish. Both the \deepbetwoknn{} variants and the \deepcetwoknn{} variants perform worse compared to the XLM-R Mix-Adapt baseline for English. Seq-Adapt is a recently published effective baseline \cite{garg2020tanda}, but for the \wul{} dataset, it does not perform well compared to the Mix-Adapt baseline. Note that the test set for the \wul{} dataset is relatively small (100 examples per language) and the examples in the test set are human translations of the English test set. Yet, we chose this dataset as it results in a larger coverage of languages. We acknowledge this limitation (that the dataset is based on translations) in our experiments and that is why we further use \jigsawm{} to demonstrate the generality of our results.

\begin{table}[tbh]
\resizebox{1.0\columnwidth}{!}{
\setlength{\extrarowheight}{3pt}
\begin{tabular}{llll}
\toprule
\bf Method              & \bf ES   & \bf IT   & \bf TR   \\
\midrule
\multicolumn{4}{c}{\emph{Fine-Tuned kNN Baselines}} \\
XLM-R Target-kNN    & 32.3 & 23.8 & 48.5 \\
XLM-R Mix-Adapt-kNN & 40.9 & 30.3 & 38.2 \\
XLM-R Seq-Adapt-kNN & 29.7 & 34.9 & 32.2 \\
\midrule
\multicolumn{4}{c}{\emph{Sentence Similarity kNN Baselines}} \\
LaBSE-kNN           & 44.7 & 48.5 & 66.0 \\
Weighted LaBSE-kNN  & 44.8 & 38.3 & 52.1 \\
\midrule
\multicolumn{4}{c}{\emph{Our Model}} \\
\small{\deepbetwoknn + \modelpxlmr $\rightarrow$ SRC}             & 59.1 & 59.5 & 81.6 \\
\small{\deepcetwoknn + \modelpxlmr $\rightarrow$ SRC}           & \textbf{61.2} & \textbf{61.1} & \textbf{85.0} \\
\midrule
\end{tabular}
}
\caption{Performance comparison in terms of F1 score for the baseline classification models and the sentence similarity model LaBSE under the majority voting \knn{} setup (experiments on Jigsaw Multilingual).}
\label{tab:knn}
\end{table}

\subsection{Impact of the Learned Voting Strategy}

To demonstrate the effectiveness of our learned voting strategy, we use our baselines (shown in Table~\ref{tab:overall}, rows 3--7) to retrieve neighbours, and then we perform majority voting to predict the label of a test instance. The results for all the approaches are shown in Table~\ref{tab:knn}. For comparison, we also add the best bi-encoder and cross-encoder versions of \knnplus (see Table~\ref{tab:overall}, rows 10 and 13).

In particular, these baseline models are pre-trained XLM-R models fine-tuned on different combinations of source and target language datasets (see \emph{Fine-Tuned \knn Baselines}, Table~\ref{tab:knn}). For each data case in the source dataset, we compute its representation as the [CLS] token from the classification model and we construct a list of vectors. Given a test data case from the target dataset, we also compute its representation based on the [CLS] token representation from the classification model. We then compute its cosine similarity with each of the [CLS] vectors from the source dataset. After that, we compute a ranked list of the top-10 neighbours based on similarity scores.

Next, we vary the number of neighbours from three to ten -- considering them in the order they are ranked based on their similarity to the query -- to obtain a majority vote and to classify the test example. We can see in Table~\ref{tab:knn} that the performance is similar to that for the LaBSE-kNN and for the Weighted LaBSE-kNN approaches in which the neighbours are retrieved using a representation space constructed from sentence similarity data (see \emph{Sentence Similarity \knn  Baselines}, Table~\ref{tab:knn}).
The results in Table~\ref{tab:knn} show that when fine-tuned models are directly used in a nearest neighbours framework without additional modifications, their performance is lower by between 25 and 60 F1 points absolute, compared to our proposed \knnplus model.

These results suggest that the interactions between the query and the retrieved neighbours captured by our model are an important prerequisite for achieving high performance.

\begin{table}[t]
\centering
\resizebox{0.80\columnwidth}{!}{
\setlength{\extrarowheight}{3pt}
    \begin{tabular}{llc}
    \toprule
    \bf{Model}                   & \bf{Representations} & \bf{F1}   \\ \midrule
    Seq-Adapt               & XLM-R            & 64.4 \\ \midrule
    \multirow{3}{*}{CE-kNN} & \modelxlmr        & 64.2 \\ 
                            & \modelpxlmr       & 62.8 \\ 
                            & \modelpxlmr $\rightarrow$ SRC    & 65.1 \\ \midrule
    \multirow{3}{*}{BE-kNN} & \modelxlmr            & 65.5 \\ 
                            & \modelpxlmr             & 63.7 \\
                            & \modelpxlmr $\rightarrow$ SRC    & \textbf{67.6} \\ \bottomrule
    \end{tabular}
}
\caption{Effectiveness of our \deepbetwoknn{} and \deepcetwoknn{} schemes in the multilingual setting that we create from \jigsawm{}.} \label{tab:multilingual}
\end{table}

\subsection{Evaluation in a Multilingual Setting}
\label{sec:multilingual_setting}

In this subsection, we go beyond our cross-lingual setting and we analyse the effectiveness of our proposed model in a multilingual setting. A multilingual setting has been explored in recent work on abusive language detection \cite{pamungkas-patti-2019-cross,ousidhoum-etal-2019-multilingual,basile-etal-2019-semeval,ranasinghe-zampieri-2020-multilingual,multilingual_hate_speech,glavas-etal-2020-xhate,leite-etal-2020-toxic} and it is desirable because online platforms are not limited to specific languages. An effective multilingual model unifies the two-stage process of language detection and prediction with a language-specific classifier. Moreover, abusive language is generally code-mixed \cite{saumya-etal-2021-offensive}, which makes language-agnostic representation spaces more desirable.

We investigate a multilingual scenario, where all target languages in our cross-lingual setting are observed both at training and at testing time. To this end, we create new training, development, and testing splits in a 5:1:2 ratio from the 8,000 available data cases in the \jigsawm dataset. Each split contains randomly sampled data in Italian, Spanish, and Turkish.

We train and evaluate our \deepbetwoknn{} and \deepcetwoknn{} using the aforementioned splits; the results are shown in Table~\ref{tab:multilingual}. Here, we must note that our neighbourhood retrieval model is language-agnostic, and thus we can retrieve neighbours for queries in any language.

\begin{table}[t]
\centering
\setlength{\tabcolsep}{3pt}
    \resizebox{1.00\columnwidth}{!}{
\begin{tabular}{lccc}
\toprule
\multicolumn{1}{c}{\bf{Text}}                 & \makecell[t]{\bf{\deepbetwoknn{}} \\ Score}  & \makecell[t]{\bf{LaBSE} \\ Score}  & \bf{Label} \\ 
\midrule
\colorbox{black}{fuck} off i do what i want                & 0.99     & 0.88        & \textcolor{flagged}{flagged}    \\
\colorbox{black}{screw} you i do what i want               & 1.0      & 0.84        & \textcolor{flagged}{flagged}    \\
i have going to do what ever i want      & -0.19    & 0.83        & \textcolor{neutral}{neutral}    \\
\colorbox{black}{fuck} u i will do as i please             & 1.0      & 0.81        & \textcolor{flagged}{flagged}    \\
\colorbox{black}{fuck} off \colorbox{black}{fuck} off i do what i want \colorbox{black}{b} & 1.0      & 0.77        & \textcolor{flagged}{flagged}    \\
nah man i do wat i want                  & -0.19    & 0.75        & \textcolor{neutral}{neutral}    \\
i shall go ahead and do it               & -0.18    & 0.74        & \textcolor{neutral}{neutral}    \\
whaaat whateva i do what i want”         & -0.2     & 0.72        & \textcolor{neutral}{neutral}    \\
ok i will do it                          & -0.17    & 0.69        & \textcolor{neutral}{neutral}    \\
great i will do what you are saying      & -0.16    & 0.68        & \textcolor{neutral}{neutral} \\  
\bottomrule
\end{tabular}
}
\caption{An example showing the effectiveness of our bi-encoder representation space for computing the similarity between the query (flagged) and its neighbours. We masked the offensive tokens in the examples for better reading experience.}
\label{tab:be_example}
\end{table}

We find that in a multilingual scenario, our \deepbetwoknn{} model with SRC pre-training performs better than the \deepcetwoknn{} model. Both the BE and the CE approaches supersede the best baseline model Seq-Adapt. Compared to the cross-lingual setting, there is more data in a mix of languages available. We hypothesise that the success of the bi-encoder model over the cross-encoder one stems from the increase in data size.

\begin{figure*}[tbh]
    \centering
    \includegraphics[width=1.0\textwidth]{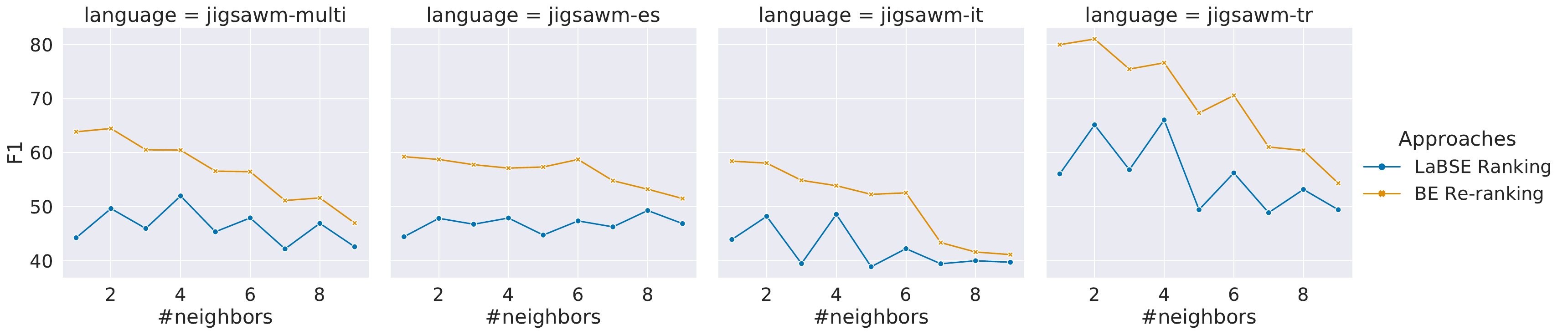}
    \caption{Impact of re-ranking neighbours using LaBSE in the \deepbetwoknn{} representation space.}
    \label{fig:reranking}
\end{figure*}

\begin{figure*}[tbh]
    \centering
    \includegraphics[width=\textwidth]{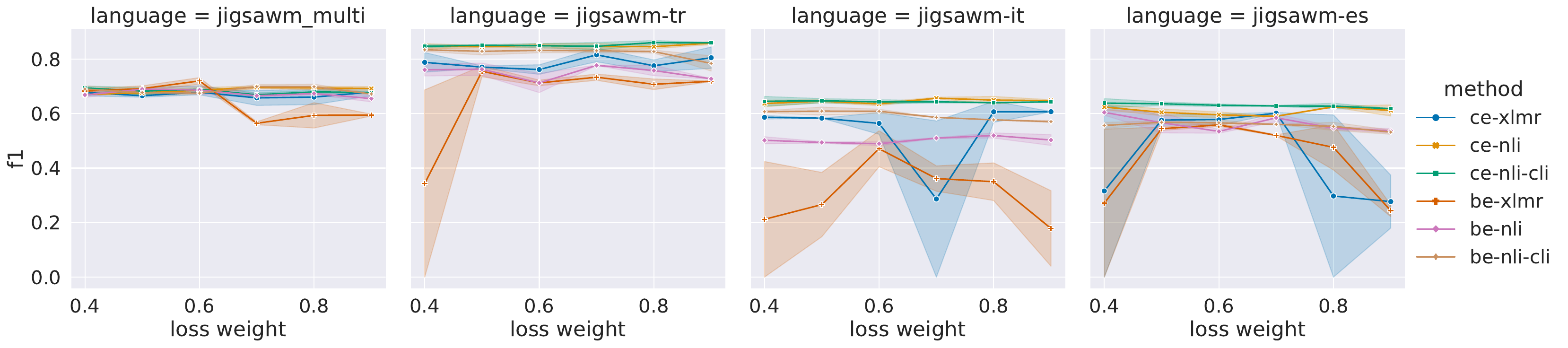}
    \caption{Multi-task loss parameter sensitivity with  uncertainties from two learning rates: 5e-05, 7e-05.}
    \label{fig:loss_sensitivity}
\end{figure*}

\subsection{Analysis of the BE Representation}

In order to understand the impact of the representations by \deepbetwoknn{} + \modelpxlmr $\rightarrow$ SRC, a model variant instantiated from our proposed \knn{}{} framework, we computed the similarity between the query and its neighbours in the representation space. An example is shown in Table~\ref{tab:be_example} (it is the example from the introduction). Given the Turkish flagged query, we use LaBSE~\cite{feng2020language} and our BE representation space to retrieve ranked lists of its ten nearest neighbours. The table shows the scores computed by both approaches, and we  can see that our representation can help discriminate between flagged and neutral contents better. When we compute the cosine similarity between the query and the nearest neighbours, the BE representation space assigns negative scores to the neutral content. The LaBSE sentence embeddings are optimised for semantic similarity, and thus using them does not allow us to discriminate between flagged and neutral content. 

We further study the impact of our representation by comparing a voting-based \knn{} on the top-10 neighbours retrieved by LaBSE vs. a re-ranking using our BE representation.

For both the LaBSE-based ranking and for our re-ranking, at each ranking point, we apply the majority voting kNN approach on the neighbourhood within that ranking point. Figure~\ref{fig:reranking} shows the results for the test part of the \jigsawm dataset (including the multilingual setup; see Section~\ref{sec:multilingual_setting}). We can see that the re-ranking step improves over LaBSE for all the different numbers of neighbours.

\subsection{Multi-Task Learning Parameter Sensitivity} 

Our approach uses multi-task learning, where we balance the weights of $\mathcal{L}_{cll}$ and $\mathcal{L}_{lal}$ using a hyper-parameter $\lambda$. Figure~\ref{fig:loss_sensitivity} shows the impact of different values for this hyper-parameter. On the horizontal axis, we increase the importance of the $\mathcal{L}_{lal}$ loss, and we show the performance of all model variants on the development part of the \jigsawm dataset. We can see that the models perform well if the weight for the label-agreement loss is set to 0.7, and degrades if it is increased.

\section{Conclusion and Future Work}

We proposed \knnplus{}, a novel framework for cross-lingual content flagging, which significantly outperforms strong baselines with limited training data in the target language. We further demonstrated the effectiveness of our framework in a multilingual scenario, where a test data point can be in Turkish, Italian, or Spanish.

Moreover, we provided a qualitative analysis of the representations learned by our proposed \deepbetwoknn{} framework, and we demonstrated that, in the learned representation space, flagged content stays close to flagged content, while non-flagged stays close to non-flagged content.

Our framework computes a neighbourhood representation for a query using an attention mechanism, thus indicating the influence of each individual neighbour. This and the \knn-based architecture offer an opportunity to obtain an explanation for the individual model predictions, and such explanations can be based not only on the textual content of the influential neighbours, but also on their original fine-grained labels.

In future work, we plan to understand the viability of such explanations in a user study.
We also plan to evaluate our framework on other content flagging tasks, e.g.,~for detecting harmful memes \cite{dimitrov-etal-2021-detecting,pramanick-etal-2021-detecting,pramanick-etal-2021-momenta-multimodal}, as the framework is not limited to abusive content detection. 

\section*{Acknowledgements}

We would like to thank the entire Checkstep team for the useful discussions on the potential implications of this research. We would especially like to thank Jay Alammar, who further provided feedback on the model and created the general conceptual diagram that explains our proposed neighbourhood framework.

\bibliography{tacl2018}
\bibliographystyle{acl_natbib}

\end{document}

%% file: macros.tex
\usepackage{soul}

\definecolor{Gray}{gray}{0.9}
\definecolor{Blue}{rgb}{0.8,0.85,1}
\definecolor{Red}{rgb}{1,0.85,0.8}
\definecolor{flagged}{RGB}{153, 0, 0}
\definecolor{neutral}{RGB}{56, 118, 29}

%% file: introduction.tex
\section{Introduction}

Online content moderation is an increasingly important problem -- small-scale websites and large-scale corporations alike strive to remove harmful content from their platforms~\cite{vidgen-etal-2019-challenges,pavlopoulos-etal-2017-deeper,wulczyn2017ex}. 
This is partly in anticipation of proposed legislation, such as the \textit{Digital Service Act} \cite{digitalservicesact2020} in the EU and the \textit{Online Harms Bill} \cite{online_harm_bill} in the UK. Moreover, the lack of content moderation can have significant impact on businesses (e.g.,~Parler was denied server space),
%\footnote{\url{https://www.nbcnews.com/tech/tech-news/amazon-suspends-hosting-parler-its}\linebreak\url{-servers-citing-violent-content-n1253648}}), 
on governments (e.g.,~the U.S. Capitol Riots),
%\footnote{\url{https://www.cbsnews.com/news/capitol-riot-arrests-2021-02-27/}}), 
and on individuals, e.g., because hate speech is linked to self-harm \cite{jurgens-etal-2019-just}.

A key challenge when developing content moderation systems is the lack of resources for many languages (other than English). With this in mind, here we aim to create a content flagging model for a target language with limited annotated data by transferring knowledge from another dataset in a different language, for which a large amount of training data is available.

Various approaches have been proposed in the literature to address the lack of enough training data in the target language. A popular approach is to fine-tune large-scale pre-trained multilingual language models such as XLM~\cite{lample2019cross}, XLM-R~\cite{conneau2020-xlm-roberta}, or mBERT~\cite{devlin2019bert} on the target dataset ~\cite{glavas-etal-2020-xhate, stappen2020cross}. In order to incorporate knowledge from the source dataset, a sequential adaptation technique can be used that first fine-tunes a multilingual language model (LM) on the source dataset, and then on the target dataset \cite{garg2020tanda}. There are also existing approaches for mixing the source and the target datasets \cite{shnarch2018will} in different proportions, followed by fine-tuning the multilingual language model on the resulting dataset. While sequential adaptation introduces the risk of forgetting the knowledge from the source dataset, such mixing methods are driven by heuristics that are effective, but not systematic. Crucially, as we argue in this paper, this is because they do not model the relationship between the source and the target datasets. Another problem arises if we consider that examples with novel labels can be added to the source dataset. This is a specifically pertinent issue for content moderation, as efforts to create new resources often lead to the introduction of new label inventories or taxonomies \cite{banko-etal-2020-unified}. In that case, model re-training becomes a requirement in order to be able to map the new label space to the output layer that is used for fine-tuning. 

\begin{figure}[t]
    \centering
    \includegraphics[width=\columnwidth]{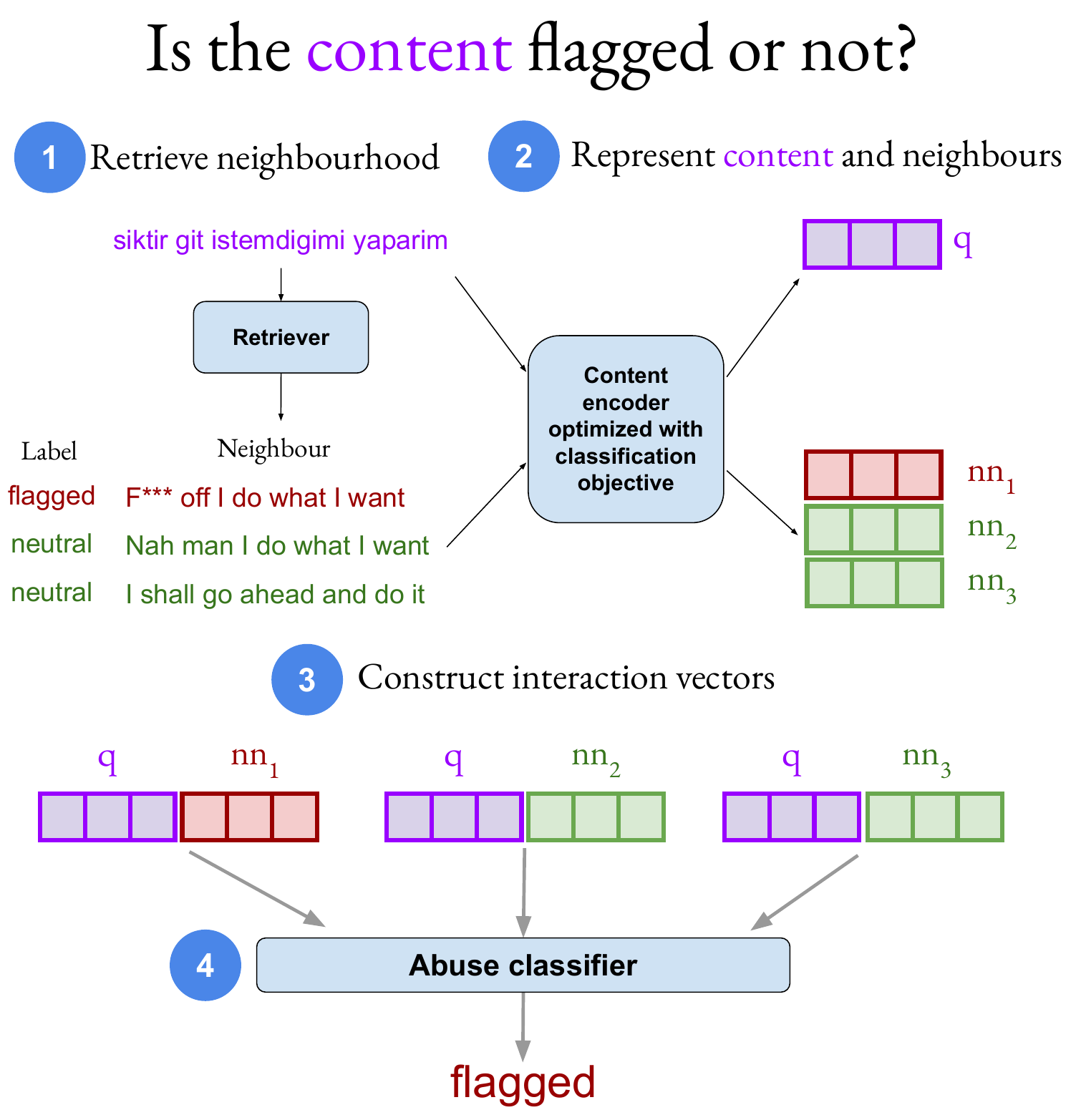}
    \caption{Conceptual diagram of our neighbourhood framework. The query is processed using run-time compute, while the neighbour vector is pre-computed.}
    \label{fig:conceptual_framework}
\end{figure}

We propose a Transformer-based $k$-Nearest Neighbour (\knnplus) framework,\footnote{We use a `+' superscript to indicate that our \knnplus framework is an improvement over the vanilla $k$-NN model.} a one-stop solution and a significant improvement over the vanilla $k$-NN model. Our framework addresses the above-mentioned challenges, which are not easy to solve via simple fine-tuning of pre-trained language models. Moreover, to the best of our knowledge, our framework is the first attempt to use $k$-NN for transfer learning for the task of abusive content detection.

Given a query, which is a training or an evaluation data point from the target dataset, \knnplus\ retrieves its nearest neighbours using a language-agnostic sentence embedding model. Then, it constructs Transformer representations for the query and for its neighbours. After that, it computes interaction features, which are based on the interactions of the representations of the query with each of its neighbours.\footnote{We borrow the terminology from information retrieval, as the interactions between a query and a document in deep matching models are computed in a similar way~\citep{drmm}.}
 At training time, the interaction features are optimised using supervised training signals computed from the label of the query and the neighbour, so that the features indicate their level of agreement. 
 
 For example, if the query and its neighbour are both abusive, they agree on the labels. Thus, the interactions help the model learn a semantic similarity space in terms of labels. The framework further uses a self-attention mechanism to aggregate the interaction features from all the neighbours, and it uses the aggregated representation to classify the input query. This representation is computed from the interaction features and indicates the agreement of the query with the neighbourhood. As the predictions are made based on aggregated interaction features only, \knnplus can easily incorporate new examples with unseen labels without requiring re-training. The conceptual framework is shown in Figure~\ref{fig:conceptual_framework}; it is robust to neighbours with incorrect labels, as it can learn to disagree with them as part of its training process.

We instantiate two variants of our framework: Cross-Encoder (CE) \knnplus and Bi-Encoder (BE) \knnplus. The \deepcetwoknn concatenates the query and a neighbour, and passes that sequence through a Transformer to obtain interaction features. \deepbetwoknn computes representations of the query and of a neighbour by passing them individually through a Transformer, and computes interaction features from these representations. \deepbetwoknn is more efficient than \deepcetwoknn, but it does not yield the same performance gains. Both models outperform six strong baselines both in cross-lingual and in multilingual settings. Our contributions can be summarised as follows:

\begin{itemize}[noitemsep]

    \item We address cross-lingual transfer learning for content flagging with limited labelled data from the target language. 
    
    \item We demonstrate that neighbourhood methods, such as \knn{} are viable candidates for approaching content flagging. 
    
    \item We propose a novel framework, \knnplus{}, which, unlike a vanilla \knn{}, models the relationship between a data point and each of its neighbours to represent the neighbourhood, using language-agnostic Transformers.
    
    \item Our evaluation results on eight languages from two different datasets for abusive language detection show sizable improvements of up to 9.5 F1 points absolute (for Italian) over strong baselines. On average, we achieve improvements of 3.6 F1 points for the three languages in the Jigsaw Multilingual dataset, and of 2.14 F1 points on the WUL dataset. 

\end{itemize}

%% file: related_work.tex
\section{Related Work}
Below, we review recent work on abusive language detection and neighbourhood approaches.

\subsection{Abusive Content Detection}

Most approaches for abusive language detection use text classification models, which have been shown to be effective for related tasks such as sentiment analysis. This includes SVMs \cite{macavaney2019hate}, CNNs \cite{georgakopoulos2018convolutional, badjatiya_www19, agrawal2018deep}, LSTMs \cite{Arango:2019, agrawal2018deep},  BiLSTMs, with attention \cite{agrawal2018deep}, Capsule networks \cite{srivastava-etal-2018-identifying}, and fine-tuned Transformers \cite{glavas-etal-2020-xhate}. All these approaches focus on single data points, while we also model their neighbourhoods. See \cite{nakov2021detecting} for a recent survey of abusive language detection.

Several papers studied the bias in hate speech detection datasets and criticised the use of within-dataset evaluations~\citep{Arango:2019, davidson, badjatiya_www19}, as this is not a realistic setting, and findings about generalisability based on such experimental settings are questionable. A more realistic and robust evaluation setting was investigated by \citet{glavas-etal-2020-xhate}, who showed the performance of online abuse detectors in a zero-shot cross-lingual setting. They fine-tuned several multilingual language models~\citep{devlin2019bert,lample2019cross,conneau2020-xlm-roberta,sanh2019distilbert,wang2020minilm} such as XLM-RoBERTa and mBERT on English datasets and observed how these models transfer to datasets in five other languages. Other cross-lingual abuse detection efforts include using Twitter user features for detecting hate speech in English, German, and Portuguese~\cite{fehn-unsvag-gamback-2018-effects}, cross-lingual embeddings~\cite{ranasinghe-zampieri-2020-multilingual}, and using multingual lexicon with deep learning~\cite{pamungkas-patti-2019-cross}. A lot of relevant research was also done as part of the OffensEval shared task at SemEval \cite{zampieri-etal-2019-predicting,zampieri-etal-2019-semeval,zampieri-etal-2020-semeval,rosenthal-etal-2021-solid}.

While understanding the performance of zero-shot cross-lingual models is interesting from a natural language understanding point of view, in reality, a platform willing to deploy an abusive language detection system can almost always provide some examples of malicious content for training. 

Thus, a few-shot or a low-shot scenario is more realistic, and we approach cross-lingual transfer learning from that perspective. We hypothesise that a nearest-neighbour model is a reasonable choice in such a scenario, and we propose several improvements over such a model.

\subsection{Neighbourhood Models}

\knn\ models have been used for a number of
NLP tasks such as part of speech tagging~\citep{daelemans-etal-1996-mbt} and morphological analysis~\citep{bosch2007efficient}, among many others. Their effectiveness is rooted in the underlying similarity function, and thus non-linear models such as neural networks can bring additional boost to their performance. More recently, \citet{DBLP:conf/iclr/KaiserNRB17} used a similarly differentiable memory that is learned and updated during training and is then applied to one-shot learning tasks.  \citet{DBLP:conf/iclr/KhandelwalLJZL20} introduced $k$-NN retrieval for improving language modelling, which \citet{kassner-schutze-2020-bert} extended to question answering (QA). \citet{guu2020realm} proposed a framework for retrieval-augmented language modelling (REALM), showing its effectiveness on three Open QA datasets. \citet{NEURIPS2020_6b493230} explored a retrieval-augmented generation for a variety of tasks, including fact-checking and QA, among others. \citet{fan-etal-2021-augmenting} introduced a $k$-NN framework for dialogue generation using pre-trained embeddings enhanced by learning an alignment function for retrieval from a set of external multi-modal evidence sources. Finally, \citet{wallace-etal-2018-interpreting} proposed a deep $k$-NN approach for interpreting the predictions from a neural network for the task of natural language inference. 

All the above approaches use neighbours as additional information sources, but do not consider the interactions between the neighbours as we do. Moreover, there is no existing work on using deep \knn\ models for cross-lingual abusive content detection.

%% file: proposed_model.tex
\section{\knnplus Framework}
\label{sec:proposed-model}

We present our \knnplus framework below.

\begin{figure*}[t!]
    \centering
    \begin{subfigure}[t]{0.55\textwidth}
    \centering
        \includegraphics[width=\textwidth]{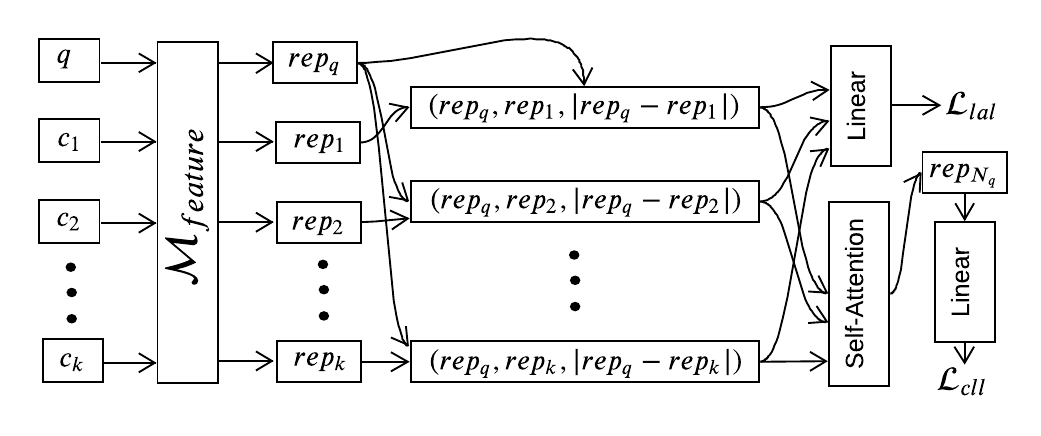}
        \caption{Bi-Encoder \knn~(\deepbetwoknn) Variant.}
        \label{fig:be_knn}
    \end{subfigure}
    \hfill
    \begin{subfigure}[t]{0.40\textwidth}
        \centering
        \includegraphics[width=\textwidth]{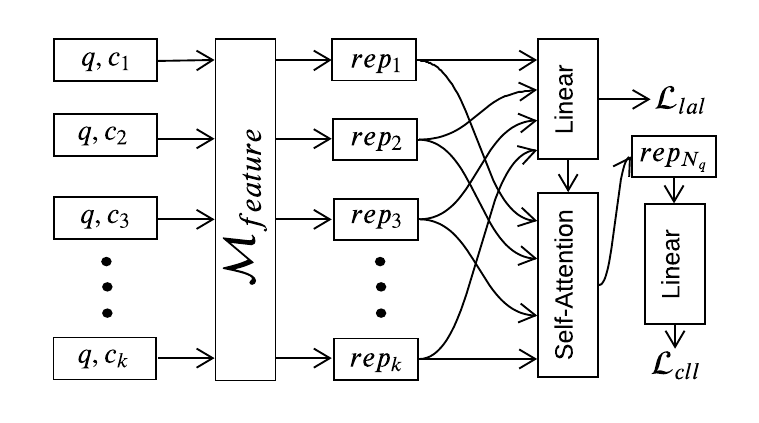}
        \caption{Cross-Encoder \knn~(\deepcetwoknn) Variant.} 
        \label{fig:ce_knn}
    \end{subfigure}
    \caption{Two variants based on two encoding schemes used in our proposed \knnplus, where $\mathcal{M}_{feature}$ is the interaction feature computation model, $q$ is the query, and $c_i$ is a candidate neighbour. In the Bi-Encoder setup (Figure~\ref{fig:be_knn}), the query and each candidate are encoded separately using the same $\mathcal{M}_{feature}$ model. Afterwards, in order to obtain a joint vector representation for each query--candidate tuple, the query's representation ($rep_{q}$) is concatenated with each candidate's representation ($rep_{i}$) along with the absolute element-wise difference between the two. In the Cross-Encoder setting (Figure~\ref{fig:ce_knn}), the query and each candidate are passed through the $\mathcal{M}_{feature}$ model, which produces the joint vector representation ($rep_{i}$) for the query--candidate tuple. Finally, we pass each joint representation through (\emph{i})~a linear layer to predict the label agreement between the query and the candidate, and (\emph{ii})~a self-attention layer followed by a linear projection layer to predict the label of the example.}
    \label{fig:knn_plus}
\end{figure*}

\subsection{Problem Setting} 
\label{sec:problem_setting}

Our goal is to learn a content flagging model from source and target datasets in different languages with different label spaces -- see Figure~\ref{fig:conceptual_framework} for an illustration of our framework.

Formally, we assume access to a source dataset for content flagging, $D^s= \big\{(x_i^s, \vb{y}_i^s)\big\}_{i=1}^{n_s}$, where $x^s_i$ is a textual content and $\vb{y}_i^s \in \mathcal{Y}$. Further, a target dataset is given, $D^t = \big\{(x_{j}^t, y_{j}^t)\big\}_{j=1}^{n_t}$, where $y_j^t \in \{flagged, neutral\}$. $D^s$ is resource-rich, i.e., $n_s \gg n_t$, and label-rich, i.e., $|   \mathcal{Y}|>2$. The label space,  $\mathcal{Y} = \{hate, insult, \ldots, neutral\}$, of $D^s$ contains fine-grained labels for different levels of abusiveness along with the \emph{neutral} label. We convert the label space of $D^s$ to align it with the label space of $D^t$ as follows: $\mathcal{Y}^\prime = \{flagged \mid x \in \mathcal{Y}, x \neq neutral\}$. Note that this conversion is needed at training time to compute label agreement in our proposed neighbourhood framework. However, 
at inference time, a conversion of the label space of $D^s$ is not needed, as the label of an item from $D^t$ is predicted using the latent representations of the neighbours, rather than their labels. This process is described in more detail in Section \ref{sec:architecture}.

\subsection{Why a Neighbourhood Framework?} 

\label{sec:motivation}
A vanilla \knn{} predicts a content label by aggregating the labels of $k$ similar training instances. To this end, it uses the content as a query to retrieve neighbours from the training instances. We hypothesise that this retrieval step can be performed in a cross-lingual transfer learning scenario. In our setting, the queries are target dataset instances, and we index the source dataset for retrieval. 

Note that the target instances could also be considered as neighbours for retrieval, but we exclude them, as the target dataset is small.   

For a vanilla \knn{}\ model, the queries and the documents are represented using lexical features, and thus the model suffers from the curse of dimensionality~\cite{knn_dimensionality}. Moreover, the prediction pipeline becomes inefficient if the source dataset is considerably larger than the target dataset, as is our case here~\cite{knn_efficiency}.
Finally, for a vanilla \knn{}, there is no straight-forward way to map between different languages for cross-lingual transfer.

We address these problems by using a Transformer-based multilingual representation space~\cite{feng2020language} that computes the similarity between two sentences expressed in different languages. We assume that efficiency issues are less critical here for two main reasons: (\emph{i})~retrieval using dense vector sentence embeddings has become significantly faster with recent advances~\cite{johnson2019billion}, and (\emph{ii})~the number of labelled source data examples is not expected to go beyond millions, because obtaining annotations for multilingual abusive content detection is costly and the annotation process can be very harmful for the human annotators as well~\cite{schmidt2017survey, waseem:2016, malmasi2018challenges, mathur2018detecting}. 

Even though multilingual language models can make the vanilla \knn{} model a viable solution for our problem, it is hard to make predictions with that model. Once a neighbourhood is retrieved, a vanilla \knn{} uses a majority voting scheme for prediction, as the example in Figure~\ref{fig:conceptual_framework} shows. Given a flagged Turkish query, our framework retrieves two \emph{neutral} and one \emph{flagged} English neighbours. Here, the majority voting prediction based on the neighbourhood is incorrect. The problem is this: \emph{A non-parametric vanilla \knn{} cannot make a correct prediction with an incorrectly retrieved neighbourhood}. Thus, we propose a learned voting strategy to alleviate this problem.

\subsection{The Architecture of \knnplus}\label{sec:architecture}

We describe our \knnplus framework (shown in Figure~\ref{fig:knn_plus}), including the training and the inference procedures. The framework includes neighbourhood retrieval, interaction feature computation and aggregation, and a multi-task learning objective function for optimisation, which we describe in detail below.  

\paragraph{Neighbourhood Retrieval}
We construct a retrieval index $R$ from the given source dataset, $D^s = \big\{(x_i^s, \vb{y}_i^s)\big\}_{i=1}^{n_s}$. For each given example $x^s_i \in D^s$, we compute its dense vector representation, $\vb{x}^s_i = \mathcal{M}_{retriever}(x^s_i)$. Here, $\mathcal{M}_{retriever}$ is a multilingual sentence embedding model that we use for retrieval. There are several multilingual sentence embedding models that we could use as $\mathcal{M}_{retriever}$~\cite{10.1162/tacl_a_00288,reimers-gurevych-2020-making, chidambaram-etal-2019-learning,feng2020language}. In this work, we use LaBSE~\cite{feng2020language}, a strong multilingual sentence matching model, which has been trained with parallel sentence pairs from 109 languages. The model is trained on 17 billion monolingual sentences and 6 billion bilingual sentence pairs and it has achieved state-of-the-art performance for a parallel text retrieval task proposed by \citet{zweigenbaum-etal-2017-overview}.
We use $\vb{x}^s_i$ as a key, and we assign $(x_i^s, \vb{y_i}^s)$ as its corresponding value. Our retrieval index $R$ stores all the key-value pairs computed from the source dataset. 

Assume we have a training data point, $(x^t_j, y^t_j)  \in D^t$, from the target dataset. We consider the content $x^t_j$ as our query $q$, i.e., $q=x^t_j$. We compute a vector representation of the query,  $\vb{q} = \mathcal{M}_{retriever}(q)$. We use $\vb{q}$ to score each key, $\vb{x}^s_i$ of $R$ using cosine similarity, i.e.,~$cos(\vb{q}, \vb{x}^s_i)$.

We sort the items in $R$ in descending order of the scores of the keys, and we take the values of the top-$k$ items to construct the neighbourhood of $q$, $N_{q} = \{(c_1, l_1), (c_2, l_2), \ldots , (c_k, l_k)\}$. Thus, each neighbour is a tuple of a content and its label from the source dataset. We convert fine-grained neighbour labels to binary labels (\emph{flagged, neutral}) as described in Section~\ref{sec:problem_setting}, to align the label space with the target dataset. Nevertheless, the original fine-grained labels of the neighbours can be used to get an explanation at inference time as this is one of the core features of \knn-based models. However, our focus is on combining these models with Transformer-based ones. We leave the investigation of the explainability characteristics of \knnplus for future work.

\paragraph{Interaction Feature Modelling} 
As discussed in Section~\ref{sec:motivation}, the neighbourhood retrieval process might lead to prediction errors. Thus, we propose a learned voting strategy to mitigate this. Our proposed strategy depends on how $q$ relates to its neighbourhood $N_q$. To model this relationship, we compute the interaction features between $q$ and the content of its $j$-th neighbour, $c_j \in N_q$. We obtain a set of $k$ interaction features from $k$ neighbours, and we optimise them using query and neighbour labels.

Similarly to \citet{reimers-gurevych-2019-sentence}, we apply two encoding schemes to compute the interaction features: a \textbf{Cross-Encoder (CE)} and \textbf{Bi-Encoder (BE)}. Under our \knnplus framework, we refer to the schemes as \deepcetwoknn for CE, and \deepbetwoknn for BE. The \deepbetwoknn is computationally inexpensive, while the \deepcetwoknn is more effective. We provide a justification for this as we describe the schemes in the following paragraphs.

For the \deepcetwoknn implementation (see Figure~\ref{fig:ce_knn}), we first form a set of query--neighbour pairs $S_{ce} = \{(q, c_1), (q, c_2), \ldots, (q, c_k)\}$ by concatenating $q$ with the content of each of its neighbours. Then, we obtain the output representation, $rep_j = \mathcal{M}_{feature}(q, c_j)$ of each $(q, c_j) \in S_{ce}$, from a pre-trained multilingual language model $\mathcal{M}_{feature}$. In this way, we create a set of interaction features, $I_{ce} = \{rep_1, rep_2, \ldots, rep_j\}$ from $q$ and its neighbourhood. Throughout this paper, the [CLS] token representation of $\mathcal{M}_{feature}$ is taken as its final output. We use varieties of implementations of $\mathcal{M}_{feature}$ in the experimentation. Figure~\ref{fig:ce_knn} shows how the interaction features are computed and optimised with a \deepcetwoknn.

Note that the feature interaction model $\mathcal{M}_{feature}$ is different from the neighbourhood retrieval one $\mathcal{M}_{retriever}$. We optimise interaction features from $\mathcal{M}_{feature}$, and we leave retrieval model optimisation for future work. 

For the \deepbetwoknn scheme (see Figure~\ref{fig:be_knn}), we obtain the output representations of $q$ and each of the neighbours individually from $\mathcal{M}_{feature}$. Given the representation of the query, $rep_q = \mathcal{M}_{feature}(q)$, and the representation of its $j^{th}$ neighbour, $rep_j = \mathcal{M}_{feature}(c_j)$, we model their interaction features by concatenating them along with their vector difference. The interaction features obtained for the $j$-th neighbour are $(rep_{q}, rep_j, |rep_{q} - rep_j|) $, and we construct a set of interaction features $I_{be}$ from all the neighbours of $q$. We use the vector difference $|rep_{q} - rep_j|$ along with the content vectors $rep_{q}$ and $rep_j$ following the work of \citet{reimers-gurevych-2019-sentence}. They trained a sentence embedding model using a Siamese neural network architecture with Natural Language Inference (NLI) data. They tried the following approaches to obtain features between the representations $u$ and $v$ of two sentences: $(u, v), (|u-v|), (u*v), (|u-v|, u*v), (u,v,u*v), (u,v,|u-v|), (u,v,|u-v|), (u*v)$. Their empirical analysis showed that $(u,v,|u-v|)$ works the best for NLI data, and thus we apply this in our framework. We plan to explore other options in future work.

Both the cross-encoder and the bi-encoder architectures were shown to be effective in a wide variety of tasks including Semantic Textual Similarity and Natural Language Inference. \citet{reimers-gurevych-2019-sentence} showed that a bi-encoder is much more efficient than a cross-encoder, and that bi-encoder representations can be stored as sentence vectors. Thus, once $\mathcal{M}_{feature}$ is trained, the vector representations $\mathcal{M}_{feature}(x_i^s)$ of each $x_i^s \in D^s$ can be saved along with the textual contents and label. Then, at inference time, only the representation of the query needs to be computed, which reduces the computation time from $k \times \mathcal{M}_{feature}$ to a constant time. Moreover, the model can easily adapt to new neighbours without the need for retraining. However, from an effectiveness perspective, the cross-encoder is usually a better option as it encodes the query and its neighbour jointly, thus enabling multi-head attention-based interactions among the tokens of the query and of the neighbour.

\paragraph{Choice of $\mathcal{M}_{feature}$}

We explore two $\mathcal{M}_{feature}$ models for both the CE and the BE schemes: a pre-trained XLM-R model, which we will refer to as \modelxlmr, as well as an XLM-R model augmented with \emph{paraphrase} knowledge, which we will refer to as \modelpxlmr~\cite{reimers-gurevych-2020-making}. Sentence representations from XLM-R are not aligned across languages~\cite{ethayarajh-2019-contextual} and \modelpxlmr overcomes this problem. In particular, \modelpxlmr is trained to learn sentence semantics with parallel data from 50 languages. Moreover, the training process includes knowledge distillation from a Sentence BERT model~\cite{reimers-gurevych-2019-sentence} trained on 50 million English paraphrases. As such, we expect \modelpxlmr to outperform \modelxlmr, as it more accurately captures the semantics of the query and its neighbour sentences.
Note that there is work on producing better alignments of multilingual vector spaces \cite{zhao2020inducing}, which would allow us to consider a variety of pre-trained sentence representation models, but exploring this is outside the scope of this paper.

\paragraph{Interaction Features Optimisation} Given a query $q$ and its $j$-th neighbour, we obtain features $rep_j \in I_{ce}$ and $(rep_{q}, rep_j, |rep_{q} - rep_j|) \in I_{be}$ from $\mathcal{M}_{feature}$ for the \deepcetwoknn and \deepbetwoknn schemes, respectively. For both schemes, we optimise the interaction features to indicate whether a query and its neighbour have the same or different labels. We do this to later aggregate interaction features from all the neighbours of a query to model the overall agreement of the query with the retrieved neighbourhood. Our hypothesis is that understanding  individual neighbour-level agreement and aggregating it will allow us also to understand the neighbourhood.

We apply a fully connected layer with two outputs over the interaction features to optimise them. The outputs indicate the label agreement between $q$ and its $j$-th neighbour, $(c_j, l_j) \in N_q$. There is a label agreement if both $q$ and the $j$-th neighbour are flagged or are both neutral, i.e.,~$y_j^t = l_j$. We learn the label agreement using a binary cross-entropy loss $\mathcal{L}_{lal}$, which is computed using the output of a softmax layer for each example in a batch of training data. We refer to $\mathcal{L}_{lal}$ as label-agreement loss. In our implementation, a batch of data comprises a query and its $k$ neighbours. We provide more details about the training procedure in Section~\ref{sec:finetune}.

Note that as our model predicts label agreement, it also indirectly predicts the label of the query and of the neighbour. In this way, it learns representations that separate flagged from the non-flagged examples. 

\paragraph{Interaction Features Aggregation}
The main reasons to use interaction features for label agreement is to predict whether $q$ should be flagged or not. In a vanilla \knn setup, there is no mechanism to back-propagate classification errors, as the only parameter to tune there is the hyper-parameter $k$. In our model, we propose to optimise the interaction features -- using a self-attention module -- to minimise the classification error with a fixed neighbourhood size $k$. To this end, we propose to aggregate the $k$ interaction features: $I_{ce}$ for \deepcetwoknn and $I_{be}$ for \deepbetwoknn. The aggregated representation captures global information, i.e.,~the agreement between the query and its neighbourhood, whereas the interaction features capture them locally. 

We use structured self-attention~\cite{lin2017structured} to capture the neighbourhood information. At first, we construct an interaction features matrix, $H \in \mathbb{R}^{k\times h}$ from the set of $k$ neighbours ($I_{ce}$ or $I_{be}$), where $h$ is the dimensionality of the interaction feature space. Then, we compute structured self-attention as follows:
\begin{align}
\vec{a} &=\operatorname{softmax}\left(W_{2} \tanh \left(W_{1} \mathbf{H}^T\right)\right)\\
			\mathbf{rep_i}&=\vec{a} \mathbf{H}
\end{align}%

Here, $W_{1} \in \mathbb{R}^{h_r \times h}$ is a matrix that encodes interactions between the representations and projects the interaction features into a lower-dimensional space, $h_r < h$, thus making the representation matrix $h_r \times k$ dimensional. We multiply another matrix $W_{2} \in \mathbb{R}^{1 \times h_r}$ by the resulting representation, and we apply softmax to obtain a probability distribution over the $k$ neighbours. Then, we use this probability distribution to produce an attention vector that linearly combines the interaction features to generate the neighbourhood representation $rep_{N_q}$, which we eventually use for classification. 

\paragraph{Classification Loss Optimisation} The aggregated interaction features, $rep_{N_q}$, are used as an input to a softmax layer with two outputs (\emph{flagged} or \emph{neutral}), which we optimise using a binary cross-entropy loss, $\mathcal{L}_{cll}$. We refer to $\mathcal{L}_{cll}$ as classification loss.

Optimising this loss means that the classification decision for a query is made by computing its agreement or disagreement with the neighbourhood as a whole. Our approach is a multi-task learning one, and the final loss is computed as follows: 
\begin{align}
\mathcal{L} = (1-\lambda) \times \mathcal{L}_{lal} + \lambda \times  \mathcal{L}_{cll}
\end{align}%

As both the classification and the label-agreement tasks aid each other, we adopt a multi-task learning approach. We balance the two losses using the hyper-parameter $\lambda$. The classification loss forces the model to predict a label for the query. As the model learns to predict a label for a query, it becomes easier for it to reduce the label agreement loss $\mathcal{L}_{lal}$. Moreover, as the model learns to predict label agreement, it learns to compute interaction features, which represent agreement or disagreement. This, in turn, helps to optimise $\mathcal{L}_{cll}$.

Note that, at inference time, our framework requires neither the labels of the neighbours for classification, nor a heuristic-based label-aggregation scheme. The classification layer makes a prediction based on the pooled representation from the interaction features, thus removing the need for any heuristic-based voting strategy based on the labels of the neighbours. Each individual interaction feature from the query and a neighbour captures the agreement between them as we optimise the features via the $L_{lal}$ loss. The opinion of the neighbourhood is captured using an aggregation of individual interaction features -- which is different from a vanilla \knn{} -- where neighbourhood opinion is captured using an individual neighbour label. As our aggregation is performed using a self-attention mechanism, we obtain a probability distribution over the interaction features that we can use to find the neighbour that influenced the neighbourhood opinion the most. We also know both the original and the converted label of the neighbour (see Section \ref{sec:problem_setting} for further details about the label space conversion). The original label of the neighbour could help us understand the prediction behind the query better. For example, if the query is flagged and the original label of the most influential neighbour is \emph{hate}, we could infer that the query is hate speech. However, we do not explore this direction in this paper, and we leave it as a future work.

%% file: tables/overall_results.tex
\begin{table*}[t]
	\centering
	\small
	\resizebox{1.00\textwidth}{!}{
	\setlength{\extrarowheight}{3pt}
		\begin{tabular}{r|lcccccccccc}            
		    \toprule

            &  & \multicolumn{3}{c}{\bf{Jigsaw Multilingual}} & \multicolumn{6}{c}{\bf{WUL}} \\ 
            
             \cmidrule(lr){3-5} \cmidrule(lr){6-11} 
             \# & \bf{Method} & ES & IT & TR & DE & EN & HR & RU & SQ & TR \\
            
            \midrule
            {1} & {Lexicon} & {35.8} & {40.5} & {34.0} & {70.9} & {70.6} & {63.9} & {63.6} & {58.2} & {71.8} \\
            2 & FastText & 55.3 & 47.2 & 64.2 & 74.2 & 72.7 & 58.9 & 74.2 & 65.9 & 72.5 \\ 
            
            \midrule
            
            3 & XLM-R Target & \underline{63.5} & 56.4 & 80.6 & 82.1 & 75.7 & 73.2 & 76.7 & 77.3 & 78.8 \\
            
            4 & XLM-R Mix-Adapt & \bf{64.2} & 58.5 & 76.1 &
            83.2 & \bf{93.9} & 87.3 & 82.1 & 86.2 & 86.0 \\ 
            
            5 & XLM-R Seq-Adapt & 60.5 & 58.3 & 81.2 &
            83.9 & 88.0 & 80.0 & 80.0 &  86.3 & 83.5 \\
            
            \midrule
            
            6 & LaBSE-kNN  & 44.7 & 48.5 & 66.0 &
            70.8 & 77.1 & 84.1 & 79.1 & 83.1 & 75.6 \\
            
            7 & Weighted LaBSE-kNN & 44.8 & 38.3 & 52.1 &
            71.7 & 85.4 & 82.4 & 79.5 & 83.7 & 81.0 \\

            \midrule
            
            \rowcolor{Red}
            
            8 & \deepcetwoknn + \modelxlmr & 58.9 & \underline{63.8} & 78.5 &
            80.4 & 83.8 & 86.2 & 77.6 & 83.5 & 85.4  \\
            
            \rowcolor{Red}
            
            9 & \deepcetwoknn + \modelpxlmr  & 59.4 & \bf{67.0} & \underline{84.4} &
            84.8 & 88.0 & 86.3 & 83.8 & 83.0 & 86.5 \\
       
            \rowcolor{Red}
            
            10 & \deepcetwoknn + \modelpxlmr $\rightarrow$ SRC  & 61.2 & 61.1 & \bf{85.0} &
            \bf{89.5} & \underline{92.3} & \bf{90.6} & 84.9 & \underline{89.5} & \underline{87.3} \\
            
            \rowcolor{Blue}
            
            11 & \deepbetwoknn + \modelxlmr  & 52.2 & 60.3 & 75.0 &
            81.6 & 80.8 & 77.9 & 78.0 & 79.6 & 79.6 \\
            
            \rowcolor{Blue}
            
            12 & \deepbetwoknn + \modelpxlmr & 58.8 & 56.6 & 80.6 &
            83.8 & 86.9 & 82.2 & \bf{86.9} & 84.9 & 83.7\\
            
            \rowcolor{Blue}
            13 & \deepbetwoknn + \modelpxlmr $\rightarrow$ SRC & 59.1 & 59.5 & 81.6 &
            \underline{88.7}  & 90.7 & \underline{87.6} & \underline{86.3} & \bf{90.2} & \bf{88.7} \\
            
            \bottomrule
        \end{tabular}
	}
	\caption{Comparison of F1 scores for the baselines and for our model variants. \deepbetwoknn and \deepcetwoknn indicate Bi-encoder and Cross-encoder schemes, respectively. SRC indicates that the model was further pre-trained with source \jigsawen, using data from it as both query and neighbours.}
	\label{tab:overall}
\end{table*}